\documentclass[10pt,twocolumn,letterpaper]{article}

\usepackage{cvpr}
\usepackage{times}
\usepackage{epsfig}
\usepackage{graphicx}
\usepackage{amsmath}
\usepackage{amssymb}

\usepackage[pagebackref=true,breaklinks=true,letterpaper=true,colorlinks,bookmarks=false]{hyperref}

\usepackage[utf8]{inputenc} 
\usepackage[T1]{fontenc}    
\usepackage{hyperref}       
\usepackage{url}            
\usepackage{booktabs}       
\usepackage{amsfonts}       
\usepackage{microtype}      
\usepackage{epsfig}
\usepackage{graphicx}
\usepackage{amsmath}
\usepackage{amssymb}
\usepackage{algorithm}
\usepackage{algpseudocode}
\usepackage{booktabs}
\usepackage{rotating}
\usepackage{booktabs} 
\usepackage{color}

\DeclareMathOperator*{\argmax}{arg\,max}
\DeclareMathOperator*{\argmin}{arg\,min}

\newcommand{\sys}{\mbox{\sc ViSeR}}

\cvprfinalcopy 

\def\cvprPaperID{0000}

\ifcvprfinal\fi
\begin{document}

\title{\sys: Visual Self-Regularization}

\author{Hamid Izadinia\thanks{Work was done while the author was an intern at Flickr, Yahoo Research.}\\
University of Washington\\
{\tt\small izadinia@cs.uw.edu}
\and
Pierre Garrigues\\
Flickr, Yahoo Research\\
{\tt\small garp@yahoo-inc.com}
}

\maketitle

\begin{abstract}
\vspace{-0.05in}

In this work, we propose the use of large set of unlabeled images as a source of regularization data for learning robust visual representation. Given a visual model trained by a labeled dataset in a supervised fashion, we augment our training samples by incorporating large number of unlabeled data and train a semi-supervised model. We demonstrate that our proposed learning approach leverages an abundance of unlabeled images and boosts the visual recognition performance which alleviates the need to rely on large labeled datasets for learning robust representation. To increment the number of image instances needed to learn robust visual models in our approach, each labeled image propagates its label to its nearest unlabeled image instances. These retrieved unlabeled images serve as local perturbations of each labeled image to perform Visual Self-Regularization (\sys). To retrieve such visual self regularizers, we compute the cosine similarity in a semantic space defined by the penultimate layer in a fully convolutional neural network. We use the publicly available Yahoo Flickr Creative Commons 100M dataset as the source of our unlabeled image set and propose a distributed approximate nearest neighbor algorithm to make retrieval practical at that scale. Using the labeled instances and their regularizer samples we show that we significantly improve object categorization and localization performance on the MS COCO and Visual Genome datasets where objects appear in context.

\end{abstract}

\vspace{-0.05in}
\section{Introduction}
\label{sec:intro}
\vspace{-0.05in}

Image recognition has rapidly progressed in the last five years. It was shown in the ground-breaking work of \cite{krizhevsky2012imagenet} that deep convolutional neural networks (CNNs) are extremely effective at recognizing objects and images. The development of deeper neural networks with over a hundred layers has kept improving performance on the ImageNet dataset~\cite{deng2009imagenet}, and we have arguably achieved human performance on this task~\cite{russakovsky2015imagenet}. These developments have become mainstream and may lead to the perception of image recognition as a solved problem. However, image recognition remains an area of active research. ImageNet is indeed biased towards single objects appearing in the middle of the image, which is in contrast with the photos we take with our mobile phones that typically contain a range of objects that appear in context. Also, the list of object categories in ImageNet is a subset of the lexical database WordNet~\cite{miller1995wordnet}. This makes ImageNet biased towards certain categories such as breeds of dogs, and does not match the scope of more general image recognition tasks such as object detection and localization in context.
 
\begin{figure*}[t]
\centering
\includegraphics[width=.99\linewidth,height=7.8cm]{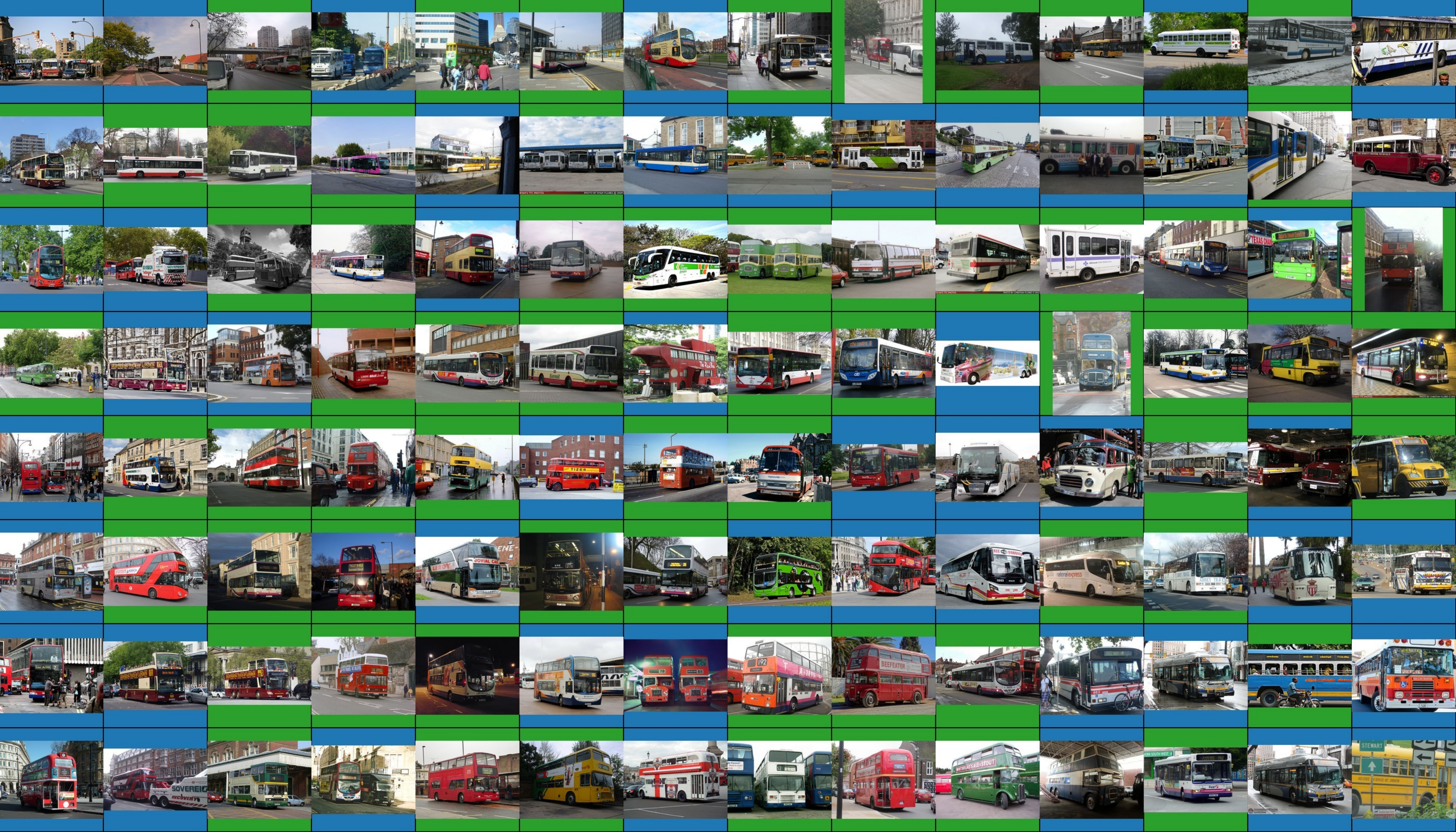}
\caption{\small
\label{fig:nearest_neighbor_map} 
The t-SNE~\cite{maaten2008visualizing} map of the whole set of images (including MS COCO and YFCC images) labeled as `Bus' category after applying our proposed \sys\ approach. Can you guess whether green or blue background correspond to the human annotated images of MS COCO dataset?}
\vspace{-0.11in}
\hfill \rotatebox[origin=c]{-180}{{\bf \small Answer key: blue: MS COCO, green: YFCC}}
\vspace{-0.15in}
\end{figure*}

Datasets such as MS COCO~\cite{lin2014microsoft} or Visual Genome~\cite{krishna2016visual} have been constructed such that photos are typically composed of multiple objects appearing at a variety of positions and scales. They provide a more realistic benchmark for image recognition systems that are intended for consumer photography products such as Flickr or Google Photos. MS COCO currently contains ~300K images and 80 object categories, whereas Visual Genome contains ~100K images and thousands of object categories. CNNs are also showing the best performance on these datasets~\cite{ren2015faster, krishna2016visual}. As training deep neural networks requires a large amount of data and the size of MS COCO and Visual Genome is an order of magnitude smaller than ImageNet, the CNN weights are initialized using the weights of a model that was originally trained on ImageNet. In this paper we focus on improving image recognition performance on MS COCO and Visual Genome. 

The labels in MS COCO and Visual Genome are obtained via crowdsourcing platforms such as Amazon Mechanical Turk. Hence it is time-consuming and expensive to obtain additional labels. However, we have access to huge quantities of unlabeled or \textit{weakly} labeled images. For example, the Yahoo Flickr Creative Commons 100M dataset (YFCC)~\cite{thomee2016yfcc100m} is comprised of a hundred million Flickr photos with user-provided annotations such as photo tags, titles, or descriptions. 

In this paper, we present a simple yet effective semi-supervised learning algorithm that is able to leverage labeled and unlabeled data to improve classification accuracy on the MS COCO and Visual Genome datasets.
We first train a fully convolutional network using the multi-labeled data (e.g. MS COCO or Visual Genome). Then, we retrieve for each training sample the nearest samples in YFCC using the cosine similarity in a semantic space of the penultimate layer in the trained fully convolutional network.
We call these \emph{Regularizer} samples which can be considered as real perturbed samples compared to the Gaussian noise perturbation considered in virtual adversarial training \cite{miyato2016distributional}. Having access to a large set of unlabeled data is critical for finding representative regularizer samples for each training instance. For making this approach practical at scale, we propose an approximate distributed algorithm to find the images with semantically similar attention activation. We then fine-tune the network using the labeled instances and \emph{Regularizer} samples. Our experimental results show that we significantly improve performance over previous methods where models are trained using only the labeled data. We also demonstrate how our approach is applicable to object-in-context retrieval.

\begin{figure*}[t]
\centering
\includegraphics[width=\linewidth]{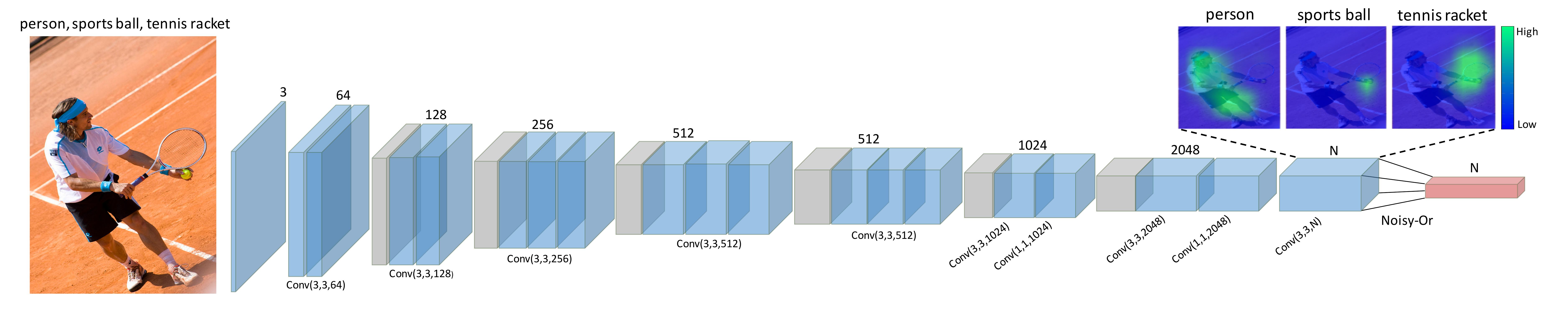}
\vspace{-0.15in}
\caption{\small
   \label{fig:net_arch}
We use a Fully Convolutional Network to simultaneously categorize images and localize the objects of interest in a single forward pass. The last layer of the network produces an tensor of N heatmaps for localizing objects where each corresponds to one of the N\emph{th} object. The green areas correspond to regions with high probability for the object produced by our network.}
\vspace{-0.05in}
\end{figure*}

\section{Related work}
\label{sec:related}
\vspace{-0.05in}

The recognition and detection of objects that appear ``in context'' is an active area of research. The most common benchmarks for this task are the PASCAL VOC ~\cite{everingham2007pascal} and MS COCO datasets. Deep convolutional neural networks have been shown to provide optimal performance in this setting with state-of-the-art performance results for object detection in~\cite{ren2015faster}. It has recently been shown in~\cite{oquab2015object,sun2015pronet,durand2017wildcat,bency2016weakly,zhou2016learning} that it is possible to accurately classify and localize objects using training data that does not contain any object bounding box information. We refer to this training data that does not contain the location information of the object as \emph{weakly-supervised}. 

The size of labeled "objects in context" datasets is typically small. For example, MS COCO has around 300,000 images and Visual Genome has over 100,000 images. However, we have access to large amounts of unlabeled web images. The Yahoo Flickr Creative Commons 100M dataset has one hundred million images that have user annotations such as tags, titles, and description. There has been some recent efforts to leverage this user annotation to build object classifiers. For instance,~\cite{izadinia2015deep} proposes a noise model that is able to better capture the uncertainty in the user annotations and improve the classification performance. It is shown in~\cite{joulin2015learning} that it is possible to learn state-of-the-art image features when training a convolutional neural network from a random initialization using user annotations as target labels. 
In~\cite{garrigues2016tag}, the authors also train deep neural networks from scratch and use the output layers as classifiers directly. However, classifier performance is lower when training on noisy data. Contrary to these approaches, we propose a form of curriculum learning~\cite{bengio2009curriculum} where we first train a model on a small set of clean data, and then augment the training set by mining instances from a large set of \emph{unlabeled} images. 

While it is shown that by making small perturbations to the input it is possible to make adversarial examples which can fool machine learning
models~\cite{szegedy2013intriguing,physicalAdversarial,AdversarialScale}, adversarial examples can be used as a means for data augmentation to improve the regularization capability of the deep models. Our method is related to adversarial training techniques~\cite{goodfellow2014explaining,miyato2016distributional,AdversarialSemiSupervised} in the sense that additional training instances with small perturbations are created and added to the training data. However, in contrast to those methods, we retrieve real adversarial examples from a large set of unlabeled images. Our examples are thus~\emph{real} image instances which possess high correlation to the labeled data in the semantic space determined by the penultimate layer of the neural network after the first phase of training. 
Such instances usually correspond to large perturbations in the input space but follow the natural distribution of the data which is analogous to the adversarial perturbations. We call our retrieved image instances as \emph{Regularizer} and show that the \emph{Regularizer} instances can be used to re-train the model and further improve performance.

Semi-supervised learning is the class of algorithms where classifiers are trained using labeled and unlabeled data. A number of approaches have been proposed in this setting such as Naive Bayes and EM algorithm~\cite{nigam2000text}, ensemble methods~\cite{bennett2002exploiting} and propagating labels based on similarity as in~\cite{zhu2002learning}. In our case the size of the unlabeled set is three orders of magnitude larger than the size of the labeled set. Existing methods are therefore impractical, and we propose a simple method to propagate labels using a nearest neighbor search. The metric is the cosine distance in the space defined by the penultimate layer of the fully convolutional neural network after it has been trained on the clean dataset. We argue that the size of our unlabeled set is critical in order for the label propagation to work effectively, and we propose approximations using MapReduce to make the search practical~\cite{dean2008mapreduce}.

Large-scale nearest neighbor search is commonly used in the computer vision community for a variety of tasks such as scene completion~\cite{hays2007scene}, image editing with the PatchMatch algorithm~\cite{barnes2009patchmatch}, or image annotation with the TagProp algorithm~\cite{guillaumin2009tagprop}.
Techniques such as TagProp~\cite{guillaumin2009tagprop} have been proposed to transfer tags from labeled to \emph{unlabeled} images. In this work we take advantage of the powerful image representation from a deep neural network to transfer labels as well as regularize training. Similarly labels can be propagated using semantic segmentation~\cite{guillaumin2014imagenet}. This method is applied on ImageNet which has a bias towards a single object appearing in the center of the image. We focus here on images where objects appear in context.

Nearest neighbor search has also been shown to be successful in other computer vision applications which involve other modalities. For example, in~\cite{devlin2015exploring} the performance of several nearest neighbor methods is examined on the image captioning task. By conducting extensive experiments, the results of~\cite{devlin2015exploring} have shown that nearest neighbor approaches can perform as good as state-of-the-art methods for image captioning.

\section{Proposed Method}
\label{sec:methods}

\subsection{Fully Convolutional Network Architecture}

Most recent developments in image recognition have been driven by optimizing performance on the ImageNet dataset. However, images in this dataset have a bias for single objects appearing in the center of the image. In order to increase performance on photos where multiple objects may appear at different scales and position, we adopt a fully convolutional neural network architecture inspired by~\cite{long2015fully}. Each fully connected layer is replaced with a convolutional layer. Hence, the output of the network is a $H \times W \times N$ tensor where the width and height  depend on the input image size and $N$ is the number of object classes. For each object class the corresponding heatmap provides information about the object's location as illustrated in Figure~\ref{fig:net_arch}. In our experiments we use the base architecture of VGG16~\cite{simonyan2014very} shown in Figure~\ref{fig:net_arch}. 

\subsection{Multiple instance learning for multilabel classification}

We are given a set of annotated images $\mathcal{A} = \{(x_i, y_i)\}_{i = 1 \ldots n}$, where $x_i$ is an image and $y_i = (y_i^{1}, \ldots, y_i^{N}) \in \{0, 1\}^N$ is a binary vector determining which object category labels are present in $x_i$. Let $f^l$ be the object heatmap for the $l$th label in the final layer of the network. The probability at location $j$ is given by applying a sigmoid unit to the logits $f^l$, e.g. $p_{j}^{l} = \sigma(f_{j}^{l})$. 

We do not have access to the location information of the objects since we are in a weakly labeled setting. Therefore, to compute the probability score for the $l$th object category to appear at the $j$th location, we incorporate a multiple instance learning approach with Noisy-OR operation~\cite{maron1998framework,zhang2005multiple,fang2015captions}. The probability for label $l$ is given by Equation~\ref{eq:lprob}. Also, for learning the parameters of the FCN, we use stochastic gradient descent to minimize the cross-entropy loss $\mathcal{L}$ formalized in Equation~\ref{eq:crossE}.

\vspace{-0.15in}
\begin{equation}
\label{eq:lprob}
p^l = 1 - \prod_j{(1 - p^l_j)}.
\end{equation}

\vspace{-0.15in}
\begin{equation}
\label{eq:crossE}
\mathcal{L} = \sum_{l=1}^{N}{-y^{l}\log{p^l} - (1 - y^{l})\log{(1 - p^l)}}
\end{equation}
\vspace{-0.1in}

\subsection{Visual Self-Regularization}

It has been observed that deep neural networks are vulnerable to adversarial examples \cite{szegedy2013intriguing}. Let $x$ be an image and $\eta$ a small perturbation such that $\|\eta\|_{\infty} \leq \epsilon$. If the perturbation is aligned with the gradient of the loss function $\eta = \epsilon sign(\nabla_x \mathcal{L})$ which is the most discriminative direction in the image space, then the output of the network may change dramatically, even though the perturbed image $\tilde{x} = x + \eta$ is virtually indistinguishable from the original. Goodfellow et al. suggest that this is due to the linear nature of deep neural networks. They also show that augmenting the training set with adversarial examples results in regularization similar to dropout\cite{goodfellow2014explaining}. 

In Virtual Adversarial Training \cite{miyato2016distributional} the perturbation is produced by maximizing 
the smoothness of the local model distribution around each data point. This method does not require the labels for data perturbations and can also be used in semi-supervised learning. The virtual adversarial example is the point in an $\epsilon$ ball around the datapoint that maximally perturbs the label distribution around that point as measured by the Kullback-Leibler divergence

\begin{figure*}[t]
\centering
\includegraphics[width=.99\linewidth,height=.73\linewidth]{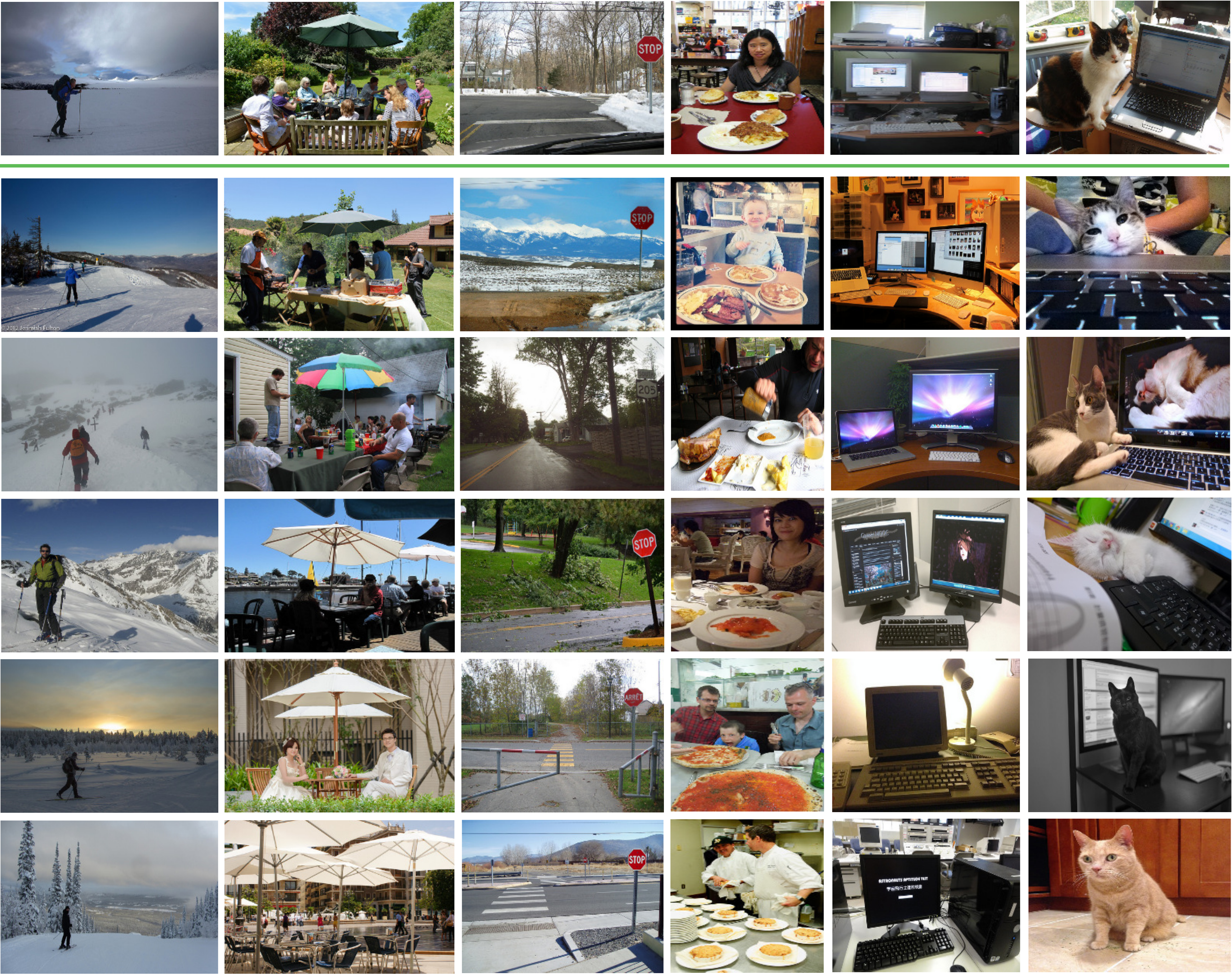}
\vspace{-0.01in}
\caption{\small
\label{fig:nearest_neighbor}
Top \emph{regularizer} examples from unlabeled YFCC dataset (row 2-6) that are retrieved for multi-label image queries in several of the MS COCO categories (first row).}
\vspace{-0.1in}
\end{figure*}

\vspace{-0.15in}
\begin{equation}
\eta = \argmin_{r: \|r\|_2 \leq \epsilon}{KL[p(y | x, \theta) ~||~ p(y | x + r, \theta)]}.
\end{equation}
\vspace{-0.1in}

We propose to draw perturbations from a large dataset of \textit{unlabeled} images $\mathcal{U}$ whose cardinality is much higher than $\mathcal{A}$. For each example $x$, we use the example $\tilde{x}$ that is nearby in the space defined by the penultimate layer in our fully convolutional network. This layer contains spatial and semantic information about the objects present in the image, and therefore $x$ and $\tilde{x}$ have similar semantics and composition while they may be far away in pixel space. We consider the cosine similarity metric to find samples which are close to each other in the feature space and for efficiency we compute the dot product of the L2 normalized feature vectors. Let $\theta$ denote the optimal parameters found after minimizing the cross-entropy loss using the training data in $\mathcal{A}$, and
$f_{\theta}(x)$ 
be the L2 normalized feature vector obtained from the penultimate layer of our network(Conv(1,1,2048)). The similarity between two images $x$ and $x'$ is then computed by their dot product $\mathcal{S}(x, x') = f_{\theta}(x)^T f_{\theta}(x')$. For each training sample $(x_i, y_i)$ in $\mathcal{A}$, we find the most similar item in $\mathcal{U}$ 

\begin{algorithm}[t]
\caption{Distributed Regularizer Sample Search}
\label{alg:mapred}
\begin{algorithmic}
\Function{Map}{$k, x$}\Comment{$k$: sample index in $\mathcal{U}$, $x$: image data}
\State Compute network output layer $f_{\theta}(x)$
\State Compute similarities with samples in $\mathcal{A}$: $s_i = f_{\theta}(x)^T f_{\theta}(x_i),  \forall i=1 \ldots n$
\State Sort $s$ by descending similarity values $s_{i_1} \geq s_{i_2} \ldots \geq s_{i_n}$
\For{$l = 1$ to $k_m$} 
	\State EMIT $i_l, (k, s_{i_l})$
\EndFor
\EndFunction

\Function{Reduce}{$k, v$}\Comment{$k$: sample index in $\mathcal{A}$, $v$: Iterator over (sample index in $\mathcal{U}$, similarity score) tuples}
\State Let $v = ((i_1, c_1), (i_2, c_2) \ldots$
\State Sort $v$ by descending similarity values $c_{j_1} \geq c_{j_2} \ldots ...$
\For{$l = 1$ to $k_r$} 
	\State EMIT $k, i_{j_l}, c_{j_l}$
\EndFor
\EndFunction

\end{algorithmic}
\end{algorithm}
\setlength{\textfloatsep}{7pt}

\vspace{-0.15in}
\begin{equation}
\tilde{x_i} = \argmax_{x \in \mathcal{U}} {\mathcal{S}(x_i, x)},
\end{equation}
\vspace{-0.15in}

\noindent and transfer the labels from $x_i$, to generate a new, \emph{Real Adversarial (Regularizer)}, training sample $(\tilde{x_i}, y_i)$. 
Similar to adversarial and virtual adversarial training, our method improves the classification performance. We interpret our sample perturbation as a form of adversarial training where additional examples are sampled from a similar semantic distribution as opposed to noise.  We also used the $\epsilon$ perturbation of each labeled sample in the gradient direction (similar to adversarial training) to find the nearest neighbor in unlabeled set and observed similar performance. Therefore, in this paper our focus is on using the labeled samples for finding \emph{Regularizer} instances to improve performance.

\subsection{Large scale approximate regularizer sample search}

In our experiments we use the YFCC dataset as our set of unlabeled images. Since it contains 100 million images, an exhaustive nearest neighbor search is impractical. Hence we use the MapReduce framework \cite{dean2008mapreduce} to find approximate nearest neighbors in a distributed fashion. Our approach is outlined in Algorithm \ref{alg:mapred}. We first pre-compute the feature representations $f_{\theta}(x_i)$ for $x_i \in \mathcal{A}$. The size of $\mathcal{A}$ for datasets such as MS COCO or Visual Genome is small enough that it is possible for each mapper to load a copy into memory. A mapper then iterates over samples $x$ in $\mathcal{U}$, computes the feature representation $f_{\theta}(x)$ and its inner product with the pre-computed features in $\mathcal{A}$. It emits tuples for the top $k_m$ matches that are keyed by the index in $\mathcal{A}$, and also contain the index in $\mathcal{U}$ and similarity score. After the shuffling phase, the reducers can select for each sample in $\mathcal{A}$ the $k_r$ closest samples in $\mathcal{U}$. We use $k_m = 1000$ and $k_r = 10$. We are able to run the search in a few hours, with the majority of the time being in the mapper phase where we compute the image feature representation. Note that our method does not guarantee that we can retrieve the nearest neighbor for each sample in $\mathcal{A}$. Indeed, if for a sample $x_i$ there exists $k_m$ samples $x_j$ such that $f_{\theta}(x_j)^T f_{\theta}(\tilde{x_i}) \geq f_{\theta}(x_i)^T f_{\theta}(\tilde{x_i})$, then the algorithm will output either no nearest neighbor or another sample in $\mathcal{U}$. However we found our approximate method to work well in practice.

\begin{figure*}[t]
\centering
\includegraphics[width=\linewidth,height=7.2cm]{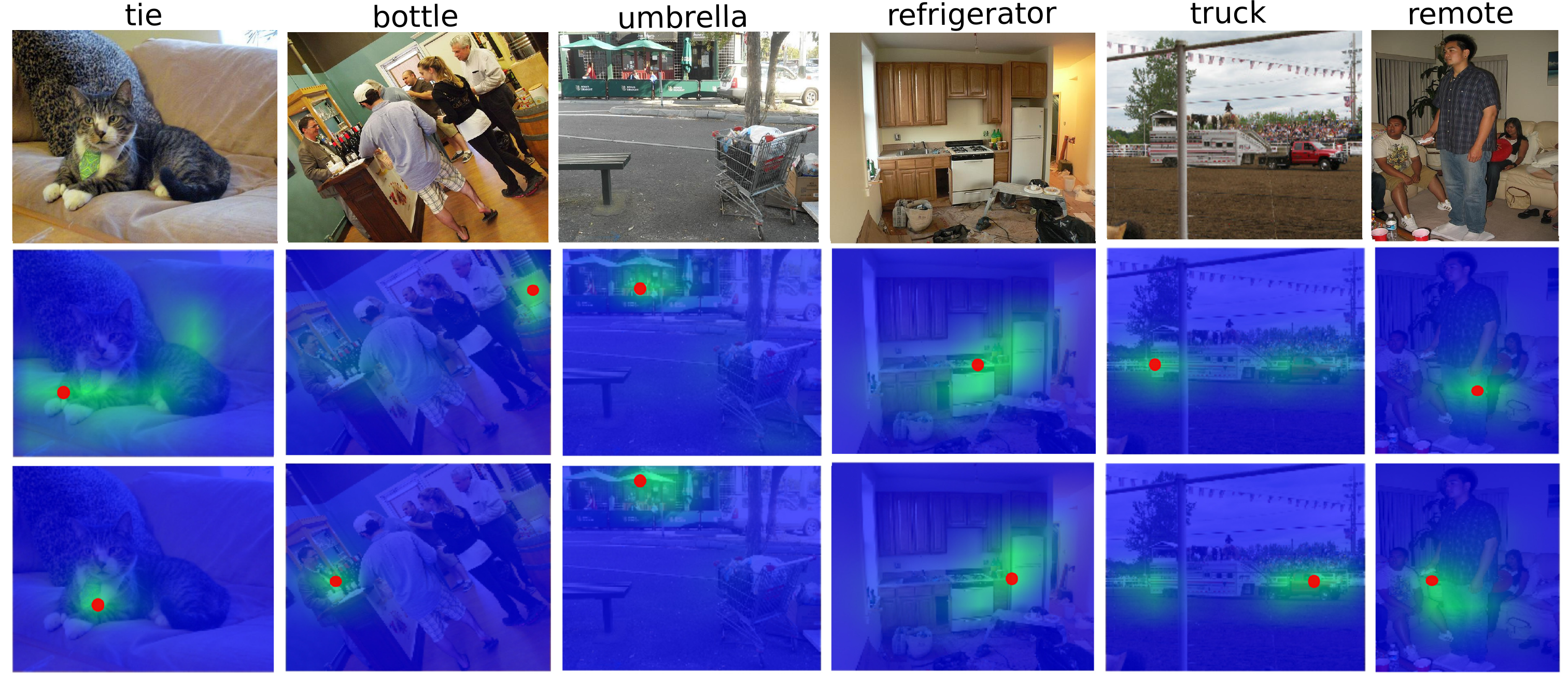}
\vspace{-0.24in}
\caption{\small
   \label{fig:FCNvsFCNaug}
Object localization comparison between ``FCN,N-OR''(mid row) and ``FCN,N-OR,\sys''(last row).}
\vspace{-0.15in}
\end{figure*}

\vspace{-0.05in}
\section{Experiments}
\label{sec:experiments}
\vspace{-0.05in}

\subsection{Semi-Supervised Multilabel Object Categorization and Localization}

We use the MS COCO~\cite{lin2014microsoft} and Visual Genome~\cite{krishna2016visual} datasets as our source of clean training data as well as for evaluating our algorithms. MS COCO has 80 object categories and is a common benchmark for evaluating object detectors and classifiers in images where objects appear in context. The more recent Visual Genome dataset has annotations for a larger number of categories than MS COCO. Applying our proposed method on the Visual Genome dataset is important to understand whether the algorithm scales to a larger number of categories, as it is ultimately important to recognize thousands of object classes in real world applications. All images for both datasets come from Flickr. In all experiments we only use the image labels for training our models and discard image captions and bounding box annotations. 

For the MS COCO dataset we use the standard split used in~\cite{lin2014microsoft} for training and evaluating the models. The training set contains 82,081 images and validation set has 40,504 images. For the Visual Genome dataset we only use object category annotations for images. The images are labeled as a positive instance for each object if the area ratio of the bounding box with regards to the image area is more than 0.025. We only consider the 1,432 object categories for which there are at least 80 image instances in the training set. 
The Visual Genome test set is the intersection of Visual Genome with the MS COCO validation set which is comprised of 17,471 images. We use the remaining 90,606 images for training our models. As for the source of unlabeled images, we use the YFCC dataset~\cite{thomee2016yfcc100m} and discard the images that are present in Visual Genome or MS COCO. 
The data is 14TB and is stored in the Hadoop Distributed File System.

We use the TensorFlow software library \cite{abadi2015tensorflow} to implement our neural networks and conduct our experiments. To conduct the distributed nearest-neighbor search, we use a CPU cluster. We use VGG16 architecture pre-trained for the ImageNet classification task as our base network. We resize images to 500$\times$500. Our initial learning rate is 0.01 and we apply the 0.1 decay factor for adapting the learning rate two times during the training after 20K and 40K mini-batches. We run stochastic gradient descent for 60K iterations with mini-batches of size 15 which corresponds to 11 epochs.

We conduct our experiments on the object classification and point-based object localization tasks. As for the object evaluation metric, we use the mean Average Precision (AP) metric where we first compute the precision for each class and then take the average over classes. For evaluating our object localization, we use the point localization metric introduced in ~\cite{oquab2015object}, where the location for a given class is given by the location with maximum response in the corresponding object heatmap.
The location is considered correct if it is located inside the bounding box associated with that class. Similar to~\cite{oquab2015object} we use a tolerance of 18 pixels.


Tables~\ref{table:coco} and ~\ref{table:vg} summarize our results with the mean AP for the classification and localization tasks on the MS COCO and Visual Genome datasets.
We compare our performance with three state-of-the-art methods for object localization and classification~\cite{oquab2015object, sun2015pronet, bency2016weakly}. In~\cite{oquab2015object}, to handle the uncertainty in object localization, the last fully connected layers of the network are considered as convolution layers and a max-pooling layer is used to hypothesize the possible location of the object in the images. In contrast, we use Noisy-OR as our pooling layer. In~\cite{sun2015pronet}, a multi-scale fully convolutional neural network called ProNet is proposed that aims to zoom into promising object-specific boxes for localization and classification. We compare against the different variants of ProNet with chain and tree cascades. Our method uses a single fully convolutional network, is simpler and has a lighter architecture as compared to ProNet. In all tables `FullyConn' refers to the standard VGG16 architecture while 'FullyConv' refers to the fully convolutional version of our network (see Figure~\ref{fig:net_arch}). The Noisy-OR loss is abbreviated as 'N-OR', and we denote our algorithm with \sys. 

We can see in Table~\ref{table:coco} that our proposed algorithm reaches $50.64\%$ accuracy in the object localization task on the MS COCO dataset which is more than a $4\%$ boost over~\cite{sun2015pronet} and a $9.5\%$ boost over~\cite{oquab2015object}. Also, without doing any regularization and by only using Noisy OR (N-OR) paired with a fully convolutional network, we obtain higher localization accuracy than Oquab et al.~\cite{oquab2015object} and different variants of ProNet~\cite{sun2015pronet}.

In the object classification task, our proposed \sys\ approach outperforms other state-of-the-art baselines of ~\cite{sun2015pronet,oquab2015object} by a margin of more than $4.5\%$ and gains an accuracy of $75.48\%$ for the MS COCO dataset.
In addition, other variants of~\cite{sun2015pronet} are less accurate than our fully convolutional network architecture with Noisy-OR pooling (`FullyConv, N-OR'). This result is consistent with the results we obtained in the object localization task. 
While the method of~\cite{bency2016weakly} obtains competitive performance on the MS COCO localization task, our method outperforms it in the classification task by a large margin of $21.4\%$. 
A recent method~\cite{durand2017wildcat} obtains classification and localization accuracy of $80.7\%$ and $53.4\%$ respectively using the deeper ResNet~\cite{he2016deep} architecture. Hence it is not directly comparable with ProNet~\cite{sun2015pronet},~\cite{bency2016weakly} and our method which use the VGG~\cite{simonyan2014very} as base network architecture.
In addition our proposed method has a label propagation step which produces a large set of labeled images with object level localization in ``object in context'' scenes and can be used in other learning methods. Also, the method of~\cite{durand2017wildcat} is based on a new pooling mechanism while our method proposes a better regularization for training ConvNets using a large scale set of unlabeled images in a semi-supervised setting and therefore is orthogonal to~\cite{durand2017wildcat}.
We also perform an ablation study and compare against other forms of regularization using our fully convolutional network architecture with Noisy-OR pooling (`FullyConv, N-OR'). 
In Table~\ref{table:coco} and ~\ref{table:vg}, we compare three forms of regularization: adversarial training (`AT')~\cite{goodfellow2014explaining}, virtual adversarial training (`VAT')~\cite{miyato2016distributional}, and our proposed \emph{Visual Self-Regularization} (\sys) using the YFCC dataset as source of unlabeled images. 

To conclude, our proposed approach outperforms state-of-the-art methods as well as several baselines by a substantial margin in object classification and localization tasks according to the results shown in Table~\ref{table:coco} and Table~\ref{table:vg}. Hence, the regularization mechanism of our proposed method results in a performance boost compared to the other forms of adversarial example data augmentation. 
We show that visual self-regularizers (\sys) make our learning robust to noise and provides better generalization capabilities.

\begin{table}[t]
\caption{\small
Mean AP for classification and localization tasks on the MS COCO dataset (higher is better).
\vspace{-0.25in}
}
\label{table:coco}
\begin{center}
\begin{tabular}{lcc}
\toprule
\multicolumn{1}{c}{Method}  &{Classification} & {Localization}\\
\midrule
Oquab et al.~\cite{oquab2015object}       & 62.8 & 41.2\\
ProNet (proposal)~\cite{sun2015pronet} & 67.8 & 43.5\\ 
ProNet (chain cascade)~\cite{sun2015pronet} & 69.2 & 45.4\\
ProNet (tree cascade)~\cite{sun2015pronet} & 70.9 & 46.4\\
Bency et al.~\cite{bency2016weakly} & 54.1 & 49.2\\
\midrule  
FullyConn & 66.68 & --\\
FullyConv,N-OR &72.52 & 47.47\\
FullyConv,N-OR,AT~\cite{goodfellow2014explaining}& 74.38 & 49.75\\
FullyConv,N-OR,VAT~\cite{miyato2016distributional} & 74.30 & 49.42\\
FullyConv,N-OR,\sys\ & {\bf 75.48} & {\bf 50.64}\\
\bottomrule
\vspace{-0.3in}
\end{tabular}
\end{center}
\end{table}


\subsection{Object-in-Context Retrieval}
To qualitatively evaluate \sys, we show several examples of the \emph{Regularizer} instances retrieved using our approach in Figure~\ref{fig:nearest_neighbor}. For each of the labeled images shown in the first row of Figure~\ref{fig:nearest_neighbor}, we show the top 5 retrieved images. As we can see, the unlabeled images retrieved by our approach have high similarity with the queried labeled image. Furthermore, most of the objects in the labeled images also appear in the retrieved images. This observation qualitatively demonstrates the effectiveness of our label propagation approach. 
It is worth mentioning that Figure~\ref{fig:nearest_neighbor} shows that the relative location of the objects in the retrieved images is fairly consistent with that of the query images. This suggests that our simultaneous categorization and localization approach can also be used for propagating bounding box annotations.

\begin{table}[t]
\caption{\small
Mean AP for classification and localization tasks on the Visual Genome dataset (higher is better). 
\vspace{-0.25in}
}
\label{table:vg}
\begin{center}
\begin{tabular}{lcc}
\toprule
\multicolumn{1}{c}{Method}  &{Classification} & {Localization}\\
\midrule
FullyConn & 9.94 & --\\
FullyConv,N-OR & 12.35 & 7.55\\
FullyConv,N-OR,AT~\cite{goodfellow2014explaining} & 13.96 & 9.05\\
FullyConv,N-OR,VAT~\cite{miyato2016distributional} & 13.95 & 9.06\\
FullyConv,N-OR,\sys\ & {\bf 14.82} & {\bf 9.74}\\
\bottomrule
\vspace{-0.3in}
\end{tabular}
\end{center}
\end{table}

Figure~\ref{fig:nearest_neighbor_map} shows the results of our \sys\  approach on the `Bus' category. We visualize the t-SNE~\cite{maaten2008visualizing} map of the whole set of images labeled as `Bus' which includes instances from both the labeled images in the MS COCO and unlabeled instances from the YFCC dataset. To produce the t-SNE visualization we take the output of the penultimate layer of our network as explained in Section~\ref{sec:methods}. We L2 normalize the feature vectors to compute the pairwise cosine similarity between images using a dot product. We visualize the t-SNE map using a grid~\cite{karpathytsne}. A different background color (blue vs. green) is assigned to images depending on whether they are from the labeled or unlabeled set. Notice that it is challenging to determine the color corresponding to each dataset as photos are from a similar domain. The images with a blue background belong to the MS COCO dataset and the images with a green background belong to the YFCC dataset. This visualization reveals that there are many images in the large unlabeled web resources that can potentially be used to populate the fully annotated dataset with more examples. This is a step forward for improving object categorization as well as decreasing human effort for supervision.

\begin{figure*}[t]
\centering
\includegraphics[width=\linewidth]{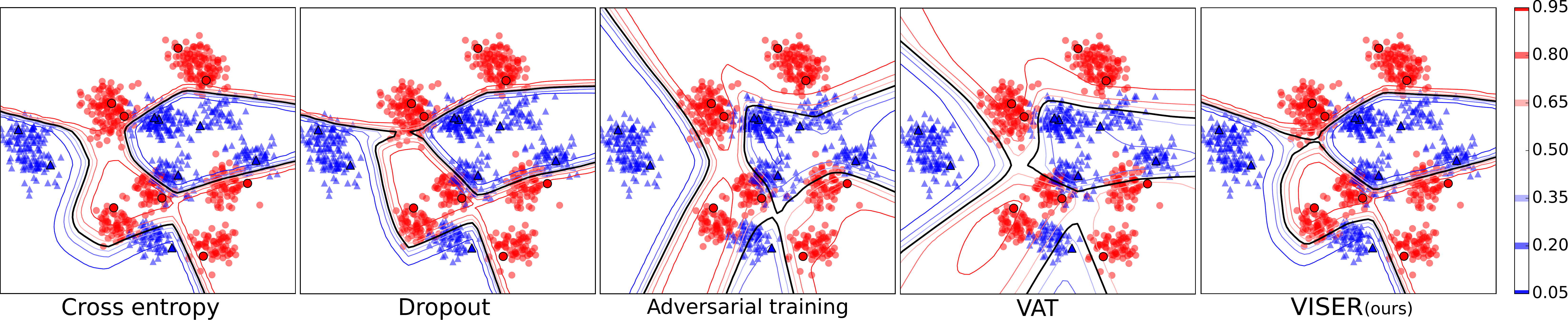}
\vspace{-0.1in}
\caption{\small
   \label{fig:synth}
Generalization comparison on a synthetic dataset between proposed \sys\ , dropout, adversarial training, and virtual adversarial training (VAT). Training samples are shown with black borders and rest of instances are test set. Each plot demonstrates the contour of the $p(y=1|x,\theta)$, from $p=0$ (blue) to $p=1$ (red).}
\vspace{-0.05in}
\end{figure*}

\begin{table*}
\caption{\small
Classification error on test synthetic dataset (lower is better).
\vspace{-0.05in}
}
\label{table:synth}
\begin{center}
\begin{tabular}{lccccc}
\toprule
\multicolumn{1}{c}{}  & cross entropy & dropout~\cite{srivastava2014dropout} & AT~\cite{goodfellow2014explaining} & VAT~\cite{miyato2016distributional} & \sys\ \\
\midrule
Error(\%) & 9.244$\pm$0.651 & 9.262$\pm$0.706 & 8.960$\pm$0.849 & 8.940$\pm$0.393 & 8.508$\pm$0.493\\
\bottomrule
\vspace{-0.45in}
\end{tabular}
\end{center}
\end{table*}

\begin{figure*}
\centering
\includegraphics[width=.98\linewidth,height=22.1cm]{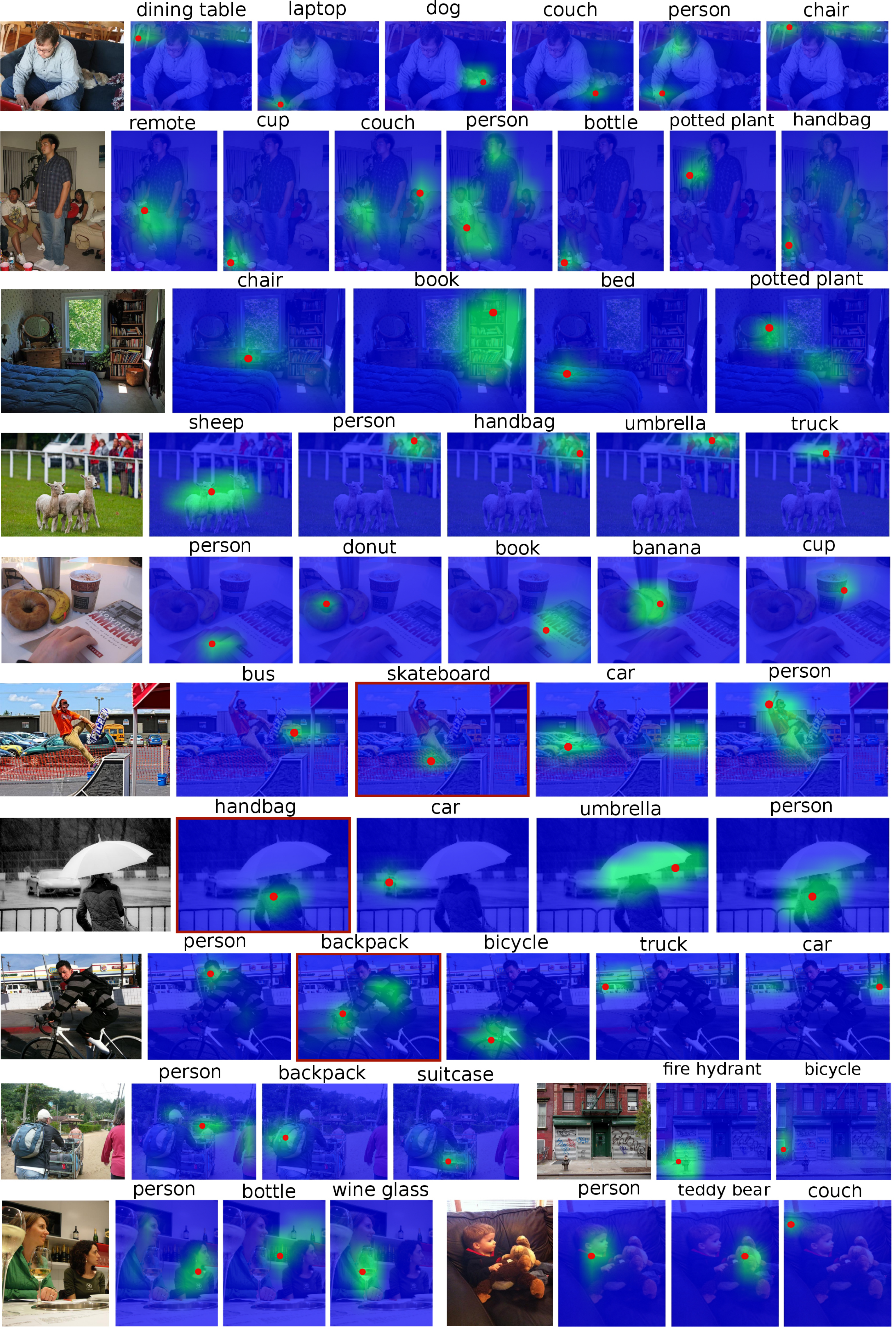}
\vspace{-0.08in}
\caption{\small
\label{fig:FCNvsFCNaug-loc}
Localization results of the proposed model on MS COCO validation set. 
Localization results of our proposed model trained on MS COCO training set and YFCC100M as the source of unlabeled set. The score map and localization of positive categories are overlaid on each image. Some failure examples are highlighted with red box for the skateboard, handbag and backpack object categories.
}
\end{figure*}

Figure~\ref{fig:FCNvsFCNaug-loc} demonstrates the qualitative performance of ``FCN,N-OR,\sys'' for multi-label localization. 
We visualize the object localization score maps where the localization regions with high probability are shown in green. We also display the localized objects using red dots. 
The score maps show that our approach can accurately localize small and big objects even in extreme cases where a big portion of the object is occluded. In the first row of Figure~\ref{fig:FCNvsFCNaug-loc} `dog' and `laptop' are localized quite accurately while they are largely occluded and truncated. 
Similarly, the third row shows the accurate localization of a `chair' although it appears in a small region of the image and is largely occluded. When there are multiple instances of an object category, such as `person' in the second row, `potted plant' in the third row, and `car' in the sixth row, all regions corresponding to these instances get a high score. 

The failure cases of our approach are distinguished via red boxes in Figure~\ref{fig:FCNvsFCNaug-loc}. For instance, the `skateboard' in row 6 is localized around the region close to the person's leg. In row 8, although the `backpack' region gets a high score map, it fails to contain the highest peak and thus the localization metric considers it as a mistake. 

We show several examples of the localization score maps produced by ``FCN,N-OR'' and ``FCN,N-OR,\sys'' in Figure~\ref{fig:FCNvsFCNaug}. By comparing the localized regions in green, we see that ``FCN,N-OR,\sys'' can locate both small and big objects more accurately. For example, in localizing small objects such as `tie', `bottle' and `remote', the peak of the localization region produced by ``FCN,N-OR'' has a large distance with the correct location of the object while ``FCN,N-OR,\sys'' localizes these objects precisely. Also, ``FCN,N-OR'' fails to be as accurate as ``FCN,N-OR,\sys'' in localizing big objects such as `umbrella' and `refrigerator'.

\vspace{-0.1in}

\subsection{Classification on Synthetic Data}
In order to evaluate the ability of our algorithm to leverage unlabeled data to regularize learning, we generate a synthetic two-class dataset with a multimodal distribution. The dataset contains 16 training instances (each class has 8 modes with random mean and covariance for each mode and 1 random sample per mode is selected), 1000 unlabeled and 1000 test samples. We linearly embed the data in 100 dimensions. 
Since the data has different modes, we can mimic the object categorization task where each object category appears in a variety of shapes and poses, each of which can be considered as a mode in the distribution.

We use a multi layer neural network with two fully connected layers of size 100, each followed by a ReLU activation and optimized via the cross-entropy loss. We compare the generalization behavior of \sys\ with the following regularization methods: dropout~\cite{srivastava2014dropout}, adversarial training~\cite{goodfellow2014explaining}, and virtual adversarial training (VAT)~\cite{miyato2016distributional}. The contour visualization of the estimated model distribution is shown in Figure~\ref{fig:synth}. We can see that both adversarial training and virtual adversarial training are vulnerable to the location of the training sample of each mode. These regularization techniques can learn a good boundary of the class when the training instance is at the center of the mode, but they over-smooth the boundary whenever the training instance is off-center. However, our proposed \sys\ sampling from unlabeled data learns a better local class distribution as adversarial samples follow the true distribution of the data and are less biased to the training instances. The dropout technique is also learning a good regularization, but it is less smooth at the boundaries of the local modes. Table~\ref{table:synth} summarizes the misclassification error on test data over 50 independent runs on the synthetic dataset.

\vspace{-0.05in}
\section{Conclusion and Future Work}
\label{sec:conclusion}
\vspace{-0.05in}
In this paper we have presented a simple yet effective method to leverage a large unlabeled dataset in addition to a small labeled dataset to train more accurate image classifiers. 
Our semi-supervised learning approach is able to find \emph{Regularizer} examples from a large unlabeled dataset.  
We have achieved significant improvements on the MS COCO and Visual Genome datasets for both the classification and localization tasks. The  performance  of  our  approach  could  be  further improved  in  future  work  by  incorporating user provided data such as `tags'.
Also, having access to a large set of unlabeled data is fairly common in other domains and hence we believe our approach could be applicable beyond visual recognition.

{\small
\bibliographystyle{ieee}
\bibliography{ViSeR_main}
}

\end{document}